\documentclass{article}

\PassOptionsToPackage{numbers, compress}{natbib}

\usepackage[final]{neurips_2024}

\usepackage{amsmath,amsfonts,bm}

\newcommand{\F}{\mF}

\newcommand{\f}{\vf}
\newcommand{\x}{\vx}
\newcommand{\y}{\vy}

\newcommand{\veps}{\bm{\varepsilon}}
\newcommand{\eps}{\bm{\epsilon}}
\newcommand{\dt}{\mathrm{d}t}
\newcommand{\dm}{\mathrm{d}}
\newcommand{\cskip}{c_{\text{skip}}}

\newcommand{\cout}{c_{\text{out}}}

\def\eqref#1{equation~\ref{#1}}

\def\1{\bm{1}}

\def\eps{{\epsilon}}

\def\vf{{\bm{f}}}
\def\vg{{\bm{g}}}

\def\vx{{\bm{x}}}
\def\vy{{\bm{y}}}

\def\mD{{\bm{D}}}

\def\mF{{\bm{F}}}
\def\mG{{\bm{G}}}

\DeclareMathAlphabet{\mathsfit}{\encodingdefault}{\sfdefault}{m}{sl}
\SetMathAlphabet{\mathsfit}{bold}{\encodingdefault}{\sfdefault}{bx}{n}

\def\gD{{\mathcal{D}}}

\def\gL{{\mathcal{L}}}

\def\gN{{\mathcal{N}}}

\newcommand{\E}{\mathbb{E}}

\usepackage[position=bottom]{caption}
\usepackage{amsmath,amsfonts,bm}
\usepackage[utf8]{inputenc} 
\usepackage[T1]{fontenc}    
\usepackage{url}            
\usepackage{booktabs}       
\usepackage{nicefrac}       
\usepackage{microtype}      
\usepackage[numbers]{natbib}
\usepackage[dvipsnames,table,xcdraw]{xcolor}
\definecolor{cvprblue}{rgb}{0.21,0.49,0.74}
\usepackage{graphicx}
\usepackage{caption}
\usepackage{arydshln}
\usepackage{tikz}
\usepackage{wrapfig}
\usepackage{float}
\usepackage{multirow}
\usepackage{pifont}
\usepackage{makecell}
\usepackage{subfigure}
\usepackage{pdfpages}
\usepackage{xcolor}
\usepackage{mathtools}

\usepackage{tikz}
\usepackage{xcolor}

\usepackage{listings}

\usepackage{ulem}

\usepackage[pagebackref,breaklinks,colorlinks,citecolor=cvprblue]{hyperref}

\usepackage{mdframed}
\definecolor{mygray}{gray}{0.97}
\colorlet{shadecolor}{mygray}
\newmdenv[%
  backgroundcolor=mygray, 
  linewidth=0pt
]{newshaded}

\newcommand{\eqnref}[1]{Eqn.~(\ref{#1})}

\title{Skywork UniPic 3.0: Unified Multi-Image Composition via Sequence Modeling}

\author{
    \textnormal{Skywork} \\
    \\
    Project Page: \url{https://skywork-unipic-v3.github.io}
}

\begin{document}

\maketitle

\begin{abstract}

The recent surge in popularity of Nano-Banana and Seedream 4.0 underscores the community's strong interest in multi-image composition tasks. Compared to single-image editing, multi-image composition presents significantly greater challenges in terms of consistency and quality, yet existing models have not disclosed specific methodological details for achieving high-quality fusion. Through statistical analysis, we identify Human-Object Interaction (HOI) as the most sought-after category by the community. We therefore systematically analyze and implement a state-of-the-art solution for multi-image composition with a primary focus on HOI-centric tasks. We present Skywork UniPic 3.0, a unified multimodal framework that integrates single-image editing and multi-image composition. Our model supports an arbitrary (1$\sim$6) number and resolution of input images, as well as arbitrary output resolutions (within a total pixel budget of 1024$\times$1024). To address the challenges of multi-image composition, we design a comprehensive data collection, filtering, and synthesis pipeline, achieving strong performance with only 700K high-quality training samples. 
Furthermore, we introduce a novel training paradigm that formulates multi-image composition as a sequence-modeling problem, transforming conditional generation into unified sequence synthesis.
To accelerate inference, we integrate trajectory mapping and distribution matching into the post-training stage, enabling the model to produce high-fidelity samples in just 8 steps and achieve a $12.5\times$ speedup over standard synthesis sampling. Skywork UniPic 3.0 achieves state-of-the-art performance on single-image editing benchmark and surpasses both Nano-Banana and Seedream 4.0 on multi-image composition benchmark, thereby validating the effectiveness of our data pipeline and training paradigm. Code, models and dataset are publicly available to support further research.

\end{abstract}

\begin{figure*}[t]
    \centering
    \includegraphics[width=\textwidth]{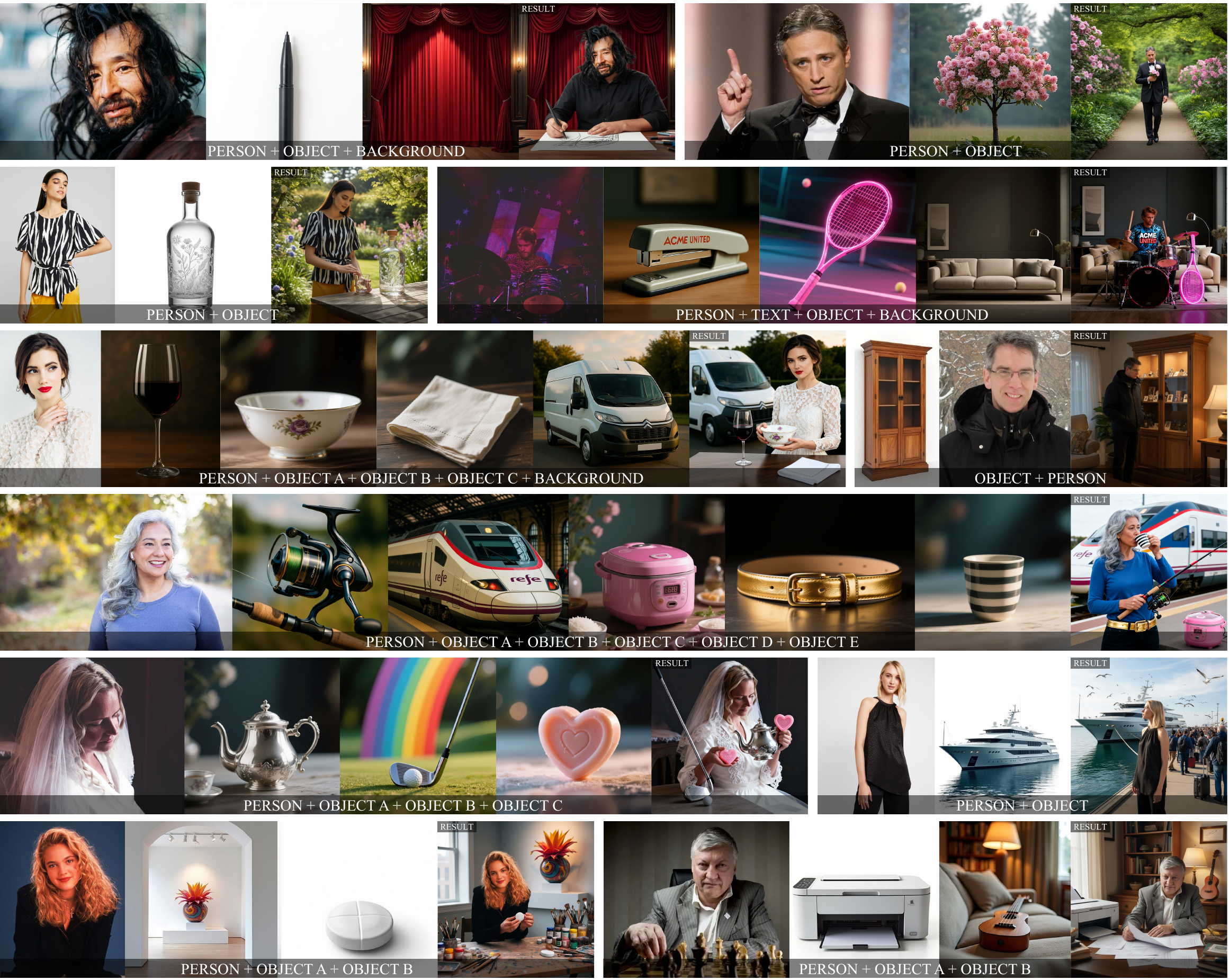}
    \caption{Our model supports image editing and composition conditioned on 1$\sim$6 input images.}
    \label{fig:teaser}
\vspace{-5mm}
\end{figure*}

\section{Introduction}
The rapid advancement of diffusion models has revolutionized generative AI, enabling unprecedented capabilities in text-to-image synthesis~\cite{flux2024,zhuo2024lumina,Cui_2024_CVPR,wei2025tiif,wang2025videoverse,du2025vqraerepresentationquantizationautoencoders,du2024aloreefficientvisualadaptation,wei2025learning,wei2025modeling,wang2025uniglyphunifiedsegmentationconditioneddiffusion} and image editing~\cite{liu2025step1x,batifol2025flux,lin2024pixwizard}. While early efforts primarily focused on generating images from scratch or modifying single images, recent community interest has shifted toward more complex scenarios involving multiple images. The viral success of systems like Nano-Banana~\cite{Nano-Banana} and Seedream 4.0~\cite{seedream2025seedream40nextgenerationmultimodal} demonstrates a growing demand for multi-image composition, where elements from different source images are seamlessly blended into a coherent, high-quality output.

Despite this surge in interest, multi-image composition remains fundamentally challenging. Unlike single-image editing, where structural consistency is largely preserved, composing multiple images requires reconciling potentially conflicting semantics, lighting conditions, perspectives, and artistic styles. According to our statistical analysis, the community shows particular interest in the category of Human-Object Interaction (HOI). HOI-centric composition demands precise spatial relationships, realistic occlusions, and natural interactions between subjects and objects. Furthermore, current approaches remain closed-source without disclosing critical implementation details, or they only support a limited number of input images and fixed resolutions during inference, such as Qwen-Image-Edit-2509~\cite{wu2025qwenimage}.

To address these limitations, we present \textbf{Skywork UniPic 3.0}, a unified framework that seamlessly integrates single-image editing and multi-image composition within a single model. Our key insight is to formulate the two tasks as conditional generation on a unified sequence representation, enabling a common architecture and training objective. This unified perspective not only simplifies model design but also facilitates knowledge transfer across tasks. 

Our proposal details and main contributions are threefold:
\begin{itemize}
    \item We propose a comprehensive data curation pipeline specifically tailored for multi-image composition. Recognizing that data quality outweighs quantity for this delicate task, we construct a high-quality dataset of 215K multi-image composition examples with a focus on challenging HOI scenarios. Our pipeline employs multi-stage filtering to ensure semantic coherence, visual compatibility, and composition quality, demonstrating that a carefully curated, moderately-sized dataset is sufficient to train a state-of-the-art model.

    \item We introduce a novel sequence modeling paradigm for multi-image composition. Specifically, we concatenate the noisy latent variables of the target output image with the latents of all reference images along the sequence dimension to form a unified long sequence. This formulation enables our model to simultaneously train on single-image editing and multi-image composition tasks while maintaining architectural simplicity. The unified sequence structure naturally accommodates variable numbers of input images and arbitrary output resolutions within a flexible pixel budget.

    \item We pioneered the integration of trajectory mapping~\cite{song2023cm,lu2024scm} and distribution matching~\cite{yin2024dmd,zhou2024sid} into the distillation of large-scale multi-image composition. The distilled model produces high-fidelity results in just 8 inference steps, achieving a remarkable 12.5$\times$ speedup over standard synthesis samplers, without sacrificing generation quality.
\end{itemize}

We extensively evaluate Skywork UniPic 3.0 on established single-image editing benchmarks and our newly proposed MultiCom-Bench for multi-image composition. Our model achieves state-of-the-art performance on ImgEdit-Bench, demonstrating the effectiveness of our unified framework, as shown in Figure~\ref{fig:teaser}. More importantly, on MultiCom-Bench, Skywork UniPic 3.0 surpasses recent strong baselines including Nano-Banana~\cite{Nano-Banana} and Seedream 4.0~\cite{seedream2025seedream40nextgenerationmultimodal}. These results validate that our careful data curation and novel training paradigm directly translate to superior composition quality.

In summary, this work presents the first systematic study of high-quality multi-image composition, provides a unified and efficient solution, and establishes a new state-of-the-art in the field. We believe our findings will benefit the community and inspire future research in unified generative frameworks.

\section{Related Work}

\paragraph{Image Editing and Multi-Image Composition.}
Text-guided image editing has witnessed remarkable progress with the advent of large-scale diffusion models. Early approaches primarily focused on single-image manipulation, employing techniques such as attention control~\cite{hertz2022prompttopromptimageeditingcross}, noise inversion~\cite{huang2024dualscheduleinversiontrainingtuningfree}, and instruct-based fine-tuning~\cite{brooks2023instructpix2pix} to alter style or semantic content while preserving the original structure. 
Subsequent works~\cite{liu2025step1x,batifol2025flux,wang2025skywork,wei2025skywork,wei2025perceive}, benefiting from text-to-image pretrained models and large-scale, high-quality editing data, have led to remarkable success in single image editing.
While effective for localized changes or global style transfers, these methods struggle with the complex synthesis required for multi-image composition, where elements from distinct reference images must be harmonized into a new, coherent context. Recently, the field has seen a paradigm shift towards more compositional generation, exemplified by proprietary systems like Nano-Banana~\cite{Nano-Banana} and Seedream 4.0~\cite{seedream2025seedream40nextgenerationmultimodal}. However, these state-of-the-art systems remain closed-source, and prior academic attempts often suffer from limited input flexibility or rigid resolution constraints (e.g., Qwen-Image-Edit-2509~\cite{wu2025qwenimage}). In contrast to these task-specific approaches, we propose a unified sequence modeling framework. By treating single-image editing and multi-image composition as instances of conditional generation within a shared manifold, Skywork UniPic 3.0 achieves superior versatility and consistency, particularly in challenging Human-Object Interaction (HOI) scenarios.

\paragraph{Image Generation with Few Steps.} Efficient diffusion sampling is typically achieved by distribution matching or trajectory mapping. Distribution matching seeks to approximate the distribution of a teacher model using a few-step student model. Methods such as ADD~\cite{sauer2024sd3turbo} employ the Jensen-Shannon divergence to minimize the divergence of two distributions, while DMD~\cite{yin2024dmd,yin2025dmd2} use the reverse Kullback-Leibler divergence to learn the teacher distribution. 
These methods can produce compelling results but fail to capture full distribution due to the mode-seeking nature of the divergences they minimize.
In contrast, trajectory mapping focuses on distilling the noise-to-data trajectory of a teacher model. Notable approaches include progressive distillation~\cite{salimans2022pd}, consistency models~\cite{song2023cm,song2023icm}, trajectory consistency models~\cite{kim2023ctm,boffi2025flowmap,wang2025tim} and rectified distillation~\cite{liu2023flow,liu2023instaflow}. sCM~\cite{lu2024scm} proposes a TrigFlow scheduler and a series techniques to simplify and stabilize continuous-time consistency training. Although trajectory mapping methods reduce the number of sampling steps, the generation quality still falls behind distribution distillation. Hybrid methods, such as Hyper-SD~\cite{ren2025hypersd}, combine both distribution matching and trajectory mapping for improved performance. We adopt the hybrid framework for few-step post-training, but propose a more principled formulation and a more efficient implementation.

\section{Preliminary}

\paragraph{Generative Models.}
Given the noise from the Gaussian distribution $\veps\sim\gN(\mathbf{0},\mathbf{I})$ and the clean data from the data distribution $\x \sim p_\text{data}(\x)$, diffusion models~\cite{song2020sde,ho2020denoising,seedream2025seedream40nextgenerationmultimodal,wang2024fitv2,wang2025native,yue2024dod,wei2025skywork} learn to map the noise distribution to the data distribution. 
Given the  time range $t\in[0,T]$, the forward process utilizes the coefficients $\alpha_t$ and $\sigma_t$, such that $\x_t=\alpha_t \x + \sigma_t \veps$. UniPic 3.0 adopts the flow matching formulation, where $\alpha_t=1-t$ and $\sigma_t=t$ are in the range $t\in[0,1]$. 
The Training objective is:
\begin{equation}
    \gL_{\theta}^{\text{FM}} = \E_{\x,\veps,t}\left\|\F_\theta(\x_t,t)-\frac{\dm\x_t}{\dt}\right\|_2^2 = \E_{\x,\veps,t}\left\|\F_\theta(\x_t,t)-(\veps-\x)\right\|_2^2,
    \label{eq:fm_objective}
\end{equation}
where $F_\theta$ is the neural network with parameters $\theta$. The sampling procedure begins with Gaussian noise $\veps$ and solves PF-ODE $\frac{\dm\x_t}{\dt}$ from $t=1$ to $t=0$ with numerical solvers, usually taking multiple numbers of function evaluations.

\paragraph{Consistency Models.}
A consistency model (CM) $\f_\theta(\x_t,t)$ learns directly the mapping from any point $\x_t$ on the trajectory to clean data $\x_0$, which is parameterized as follows:
\begin{equation}
    \f_\theta(\x_t,t)=\cskip(t)\x_t+\cout(t)\F_\theta(\x_t,t),
\end{equation}
where $\cskip(t)$ and $\cout(t)$ are time-dependent coefficients that satisfy: $\cskip(0)=1$ and $\cout(1)=0$ to ensure the boundary condition $\f_\theta(\x,0)\equiv\x$. 
Continuous-time CMs~\cite{lu2024scm,song2023cm} adopt the limiting case: $\Delta t\to0$. When choosing loss function $d(\x,\y)=\|\x-\y\|_2^2$, the continuous-time CM gradient is:
\begin{equation}
    \nabla_\theta\E_{\x,\veps,t}\left[w(t)\f_\theta^\top(\x_t,t)\frac{\dm\f_{\theta^-}(\x_t,t)}{\dt}\right],
    \label{eq:continuous_cm}
\end{equation}
where $\frac{\dm\f_{\theta^-}(\x_t,t)}{\dt}$ is the tangent of $\f_{\theta^-}$ along the trajectory of the PF-ODE. 



\paragraph{Distribution Matching Distillation.}
Unlike consistency models, which learn to map the PF-ODE trajectory, distribution matching methods aim to match the student generation distribution $p_\theta$ to the teacher generation distribution $p_\text{teacher}$. In this case, samples $\x\sim p_{\theta}$ are generated in a few-steps via $\x=\mG_\theta(\veps)$\footnote{$\mG_\theta(\veps)=\Psi(\F_\theta,\veps,N)$, where $\Psi$ is an ODE solver, $\F_\theta$ is the neural network and $N$ is the number of function evaluations. This represents the process from pure Gaussian noise $\veps$ to generated samples.}, where $\veps$ represents Gaussian noise. To minimize the difference between $p_\theta$ and $p_\text{teacher}$, the $f$-divergence~\cite{renyi1961entropy,xu2025f-divergence} is used as an optimization objective:
\begin{equation}
    \min_{\theta} [\gD_f(p_\theta \parallel p_\text{teacher})] = \min_{\theta} \left[\int_{\x\sim p_\theta(\x)}p_\text{teacher}(\x)f\left(\frac{p_\theta(\x)}{p_\text{teacher}(\x)}\right)\dm\x\right].
    \label{eq:f_divergence}
\end{equation}
The choice of function $f(\cdot)$ determines the type of divergences, such as the reverse Kullback-Leibler (KL) divergence~\cite{yin2024dmd,luo2023diff_instruct}, the Fisher divergence~\cite{zhou2024sid} and the Jensen-Shannon divergence~\cite{sauer2024sdxlturbo,sauer2024sd3turbo}.

\section{Method}

\begin{figure}
    \centering
    \includegraphics[width=1\linewidth]{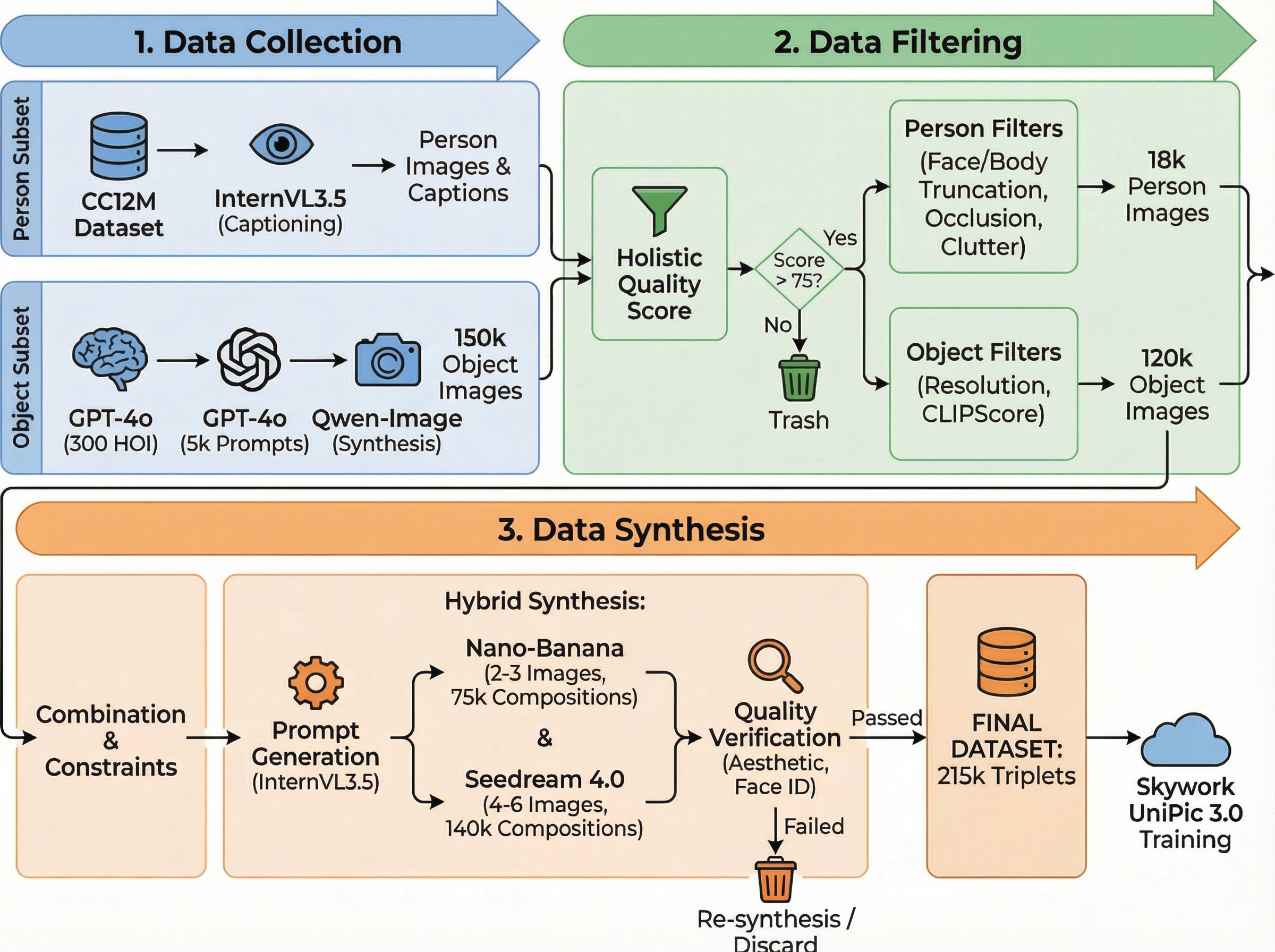}
    \caption{The overall data curation pipeline of UniPic 3.0.}
    \label{fig:data_pipeline}
    \vspace{-1.5em}
\end{figure}

\subsection{Data Curation}
To achieve remarkable multi-image composition quality for Human-Object Interaction (HOI) tasks, we meticulously design a comprehensive data curation pipeline comprising three synergistic stages: \textbf{Data Collection}, \textbf{Data Filtering}, and \textbf{Data Synthesis}. As demonstrated in Figure~\ref{fig:data_pipeline}, This pipeline yields 215K high-quality (source images, instruction, target image) triplets that serve as the foundation for training the UniPic 3.0 Model.

\paragraph{Data Collection.} We adopt a bifurcated strategy to source diverse and composition-ready person and object images. For the person subset, we curate high-quality realistic human-centric images from the CC12M~\cite{changpinyo2021conceptual} dataset, a large-scale corpus known for its rich variety of human poses, appearances, and real-world contexts. We employ InternVL3.5-38B~\cite{wang2025internvl3_5} to generate dense, structurally-aware captions that explicitly describe human attributes, clothing, pose, and scene context, which are crucial for subsequent HOI composition. For the object subset, we first prompt GPT-4o~\cite{hurst2024gpt} to generate 300 fine-grained object categories that are semantically compatible with human interaction (e.g., apparel items, handheld tools, musical instruments, furniture, and sports equipment). For each category, GPT-4o then produces 5,000 diverse textual prompts that emphasize visual attributes, materials, and typical usage contexts. These prompts are fed into Qwen-Image~\cite{wu2025qwenimage} to synthesize 150K object images. This deliberate separation ensures balanced coverage across interaction types while maintaining photorealistic quality.

\paragraph{Data Filtering.} We implement a rigorous multi-stage filtering protocol to ensure source image quality and composition suitability. First, we implement InternVL3.5-38B~\cite{wang2025internvl3_5} to assign a holistic quality score (0 to 100) to each image based on resolution, sharpness, esthetic appeal and semantic clarity, removing images that score below 75. For person images, we apply specialized face and body detectors to filter out cases with truncated heads (<90\% face visibility), occluded primary subjects (<60\% foreground occupancy), or cluttered backgrounds (background complexity score > 0.7). For object images, we enforce minimum resolution requirements (> $768^2$ pixels) and cross-check prompt-image alignment using CLIPScore~\cite{clip}, rejecting pairs with low similarity. This stringent curation retains only 18K person images and 120K object images, ensuring a clean and composition-ready inventory.

\paragraph{Data Synthesis.} The synthesis stage carefully constructs valid source image combinations and generates the corresponding target compositions. We sample 2 to 6 source images per composition while imposing hard HOI compatibility constraints through a manually curated conflict matrix: for example, a person cannot simultaneously wear two pairs of footwear or interact with two musical instruments in the same hand. This prevents physically implausible scenes. For each valid combination, we prompt InternVL3.5-38B~\cite{wang2025internvl3_5} to generate a cohesive composition prompt that describes natural spatial arrangements, realistic occlusion relationships, and harmonious scene lighting, which is a critical step that bridges the semantic gap between disparate sources. Empirically, we observe that Nano-Banana~\cite{Nano-Banana} exhibits a degraded preservation of facial identity when handling $>$ 3 input images, while Seedream 4.0~\cite{seedream2025seedream40nextgenerationmultimodal} maintains superior consistency in 4 to 6 images. Therefore, we adopt a hybrid synthesis strategy: Nano-Banana generates 75K compositions for 2 to 3 image subsets, while Seedream 4.0~\cite{seedream2025seedream40nextgenerationmultimodal} produces 140K compositions for 4 to 6 image subsets. Each synthesized target undergoes automatic quality verification via Aesthetic Score and face identity preservation checks, with failed cases re-synthesized or discarded.

\begin{figure}
    \centering
    \includegraphics[width=1\linewidth]{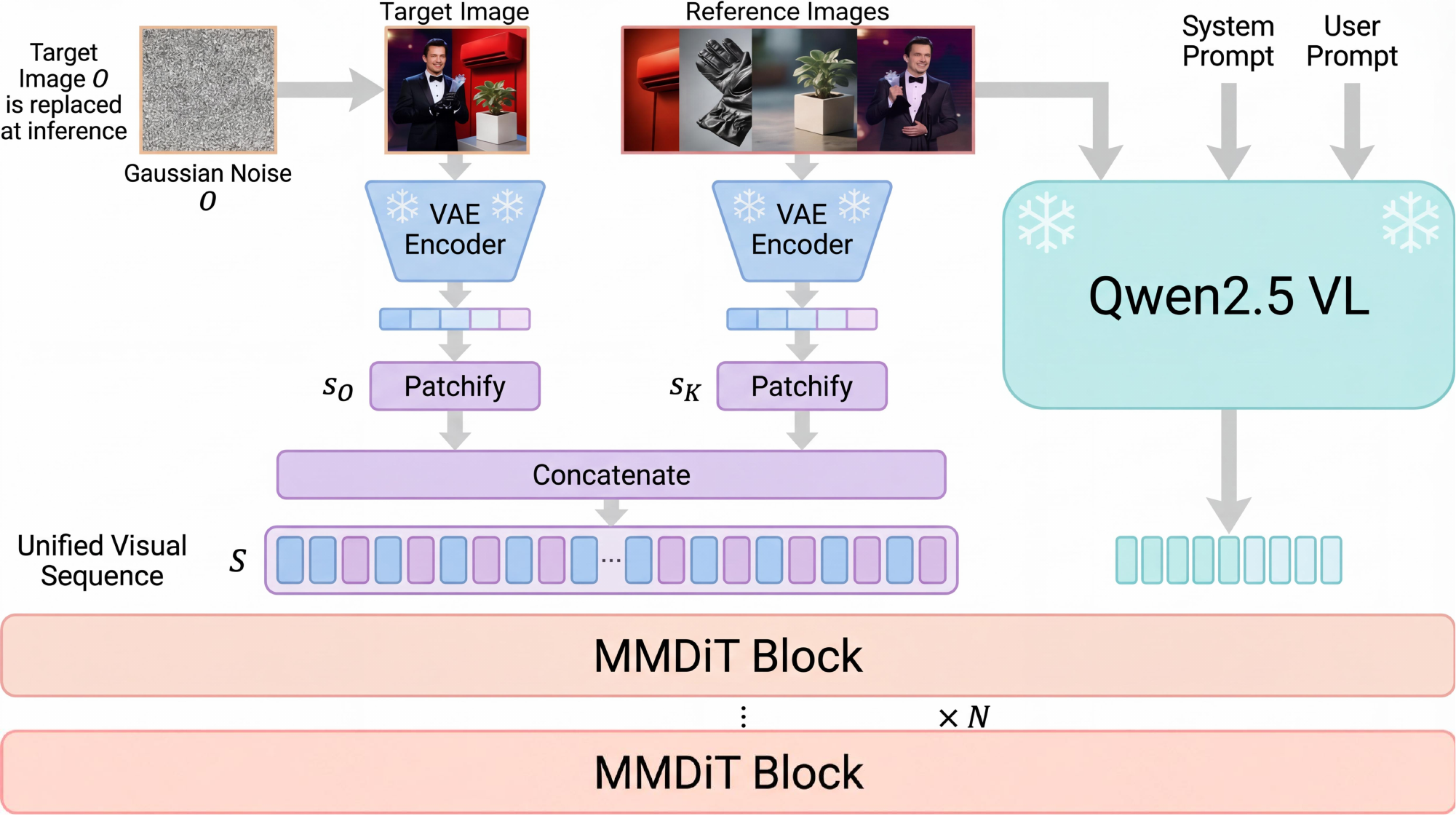}
    \caption{The overall model pipeline of UniPic 3.0.}
    \label{fig:train_pipeline}
    \vspace{-1em}
\end{figure}

\subsection{Training Paradigm}
In this section, we describe the training paradigm that enables our model to perform single image editing and multi-image composition by casting them as conditional generation over a unified visual sequence. Our model architecture follows that of Qwen-Image~\cite{wu2025qwenimage}, which incorporates Qwen2.5-VL~\cite{bai2025qwen2} as the condition encoder, employs a VAE as the image tokenizer, and utilizes MMDiT as the backbone diffusion model.

\paragraph{Latent Encoding.}
For each training instance, we sample a target image $O$ and a set of reference images $\{I_1, \dots, I_K\}$.
Each image is encoded into a latent tensor using the VAE encoder:
\begin{equation}
    z_O = f_{\text{vae-enc}}(O), 
    \quad 
    z_k = f_{\text{vae-enc}}(I_k),\; k = 1,\dots,K,
\end{equation}
where $z_O, z_k \in \mathbb{R}^{1 \times C \times H' \times W'}$, $C$ is the latent channel dimension, and $(H', W')$ is the downsampled spatial resolution.

\paragraph{Patch-wise Packing.}
Following the design of Qwen-Image, each latent tensor is reshaped into a sequence of patches by a deterministic packing operation:
\begin{equation}
    s_O = \text{pack}(z_O) \in \mathbb{R}^{N_O \times D},
    \quad
    s_k = \text{pack}(z_k) \in \mathbb{R}^{N_k \times D},
\end{equation}
where $\text{pack}(\cdot)$ rearranges $2\times2$ spatial neighborhoods into tokens, yielding a sequence length $N_{\cdot}$ and feature dimension $D$.
This operation is invertible given the spatial metadata.

\paragraph{Unified Visual Sequence.}
We construct a single unified latent sequence by concatenating the packed target and reference latents along the sequence dimension:
\begin{equation}
    S = [s_O \,\|\, s_1 \,\|\, \dots \,\|\, s_K] 
    \in \mathbb{R}^{N_{\text{tot}} \times D},
\end{equation}
where $N_{\text{tot}} = N_O + \sum_{k=1}^K N_k$.
In addition, we maintain a set of shape descriptors
\begin{equation}
    \mathcal{H} = \{h_O, h_1, \dots, h_K\},
\end{equation}
where each $h_{\cdot}$ encodes the latent height and width of the corresponding image.
These descriptors are passed into the transformer to preserve spatial structure and allow exact unpacking during reconstruction.

At inference time, we replace the true target latents with pure Gaussian noise while keeping the reference latents, as illustrated in Figure~\ref{fig:train_pipeline}.

\subsection{Post-training for Few-step Generation}


We adopt a hybrid post-training framework to convert a multi-step MMDiT model into a few-step variant. This hybrid approach unifies trajectory mapping and distribution matching, enabling generation in a few steps with high fidelity and high diversity.

\paragraph{Continuous-time Consistency Training for Flow Matching.} sCM~\cite{lu2024scm} simplifies and stabilizes continuous-time CMs using the TrigFlow formulation. To enable continuous-time consistency training for flow matching (FM) models, SANA-Sprint~\cite{chen2025sana-sprint} transform a pre-trained flow matching model into a TrigFlow model through mathematical input and output transformations. 
However, this approach changes the time variables from a standard linear schedule to a trigonometric schedule, leading to numerical instability and precision errors. 
The input and output transformations induce extra gradient terms, which affects the gradient variance and makes the training unstable.

To address these challenges, we adopt a more principled consistency flow matching training. For UniPic 3.0 transport, the consistency models are parameterized as: $\f_\theta(\x_t,t)=\x_t-t\F_\theta(\x_t,t)$, leading to the gradient:
$\nabla_\theta\f_\theta(\x_t,t) = -t\nabla_\theta\F_\theta(\x_t,t)$.
The tangent of $\f_{\theta^-}$ has the explicit form:
\begin{equation}
    \frac{\dm\f_{\theta^-}(\x_t,t)}{\dt} = \frac{\dm\x_t}{\dt} - \F_{\theta^-}(\x_t,t) - t\frac{\dm\F_{\theta^-}(\x_t,t)}{\dt} = \veps - \x - \F_{\theta^-}(\x_t,t) - t\frac{\dm\F_{\theta^-}(\x_t,t)}{\dt}.
    \label{eq:tangent_f}
\end{equation}

Taking the explicit form of \eqnref{eq:tangent_f} into \eqnref{eq:continuous_cm}, the gradient is:
\begin{equation}
    \nabla_\theta\E_{\x,\veps,t}\left[\F_\theta^\top(\x_t,t) \cdot \left(-w(t)\cdot t\cdot\frac{\dm\f_{\theta^-}(\x_t,t)}{\dt}\right)\right].
\end{equation}
Taking the identity $\nabla_\theta\E[\F_\theta^\top\y]=\frac{1}{2}\nabla_\theta\E[\|\F_\theta-\F_{\theta^-}+\y\|_2^2]$\footnote{This identity is valid for any arbitrary vector $\y$, provided that $\y$ is independent of the parameter $\theta$.} and $w(t)=\frac{1}{t}$, the consistency loss is defined as:
\begin{equation}
    \gL_\theta^\text{CM} = \left\| F_\theta(\x_t,t) - \left( \veps - \x - t\frac{\dm\F_{\theta^-}(\x_t,t)}{\dt} \right) \right\|_2^2,
    \label{eq:cm_loss}
\end{equation}
where we calculate $\frac{\dm\F_{\theta^-}(\x_t,t)}{\dt}=\frac{1}{2\epsilon}(\F_{\theta^-}(\x_{t+\epsilon},t+\epsilon) - \F_{\theta^-}(\x_{t-\epsilon},t-\epsilon))$ with the finite-difference approach~\cite{banner2007calculus,wang2025tim}, where we set $\eps=5\times10^{-3}$ by default.

\paragraph{Distribution Matching Distillation.} We adopt the reverse KL divergence to match the student distribution with the teacher distribution, leading to the gradient of \eqnref{eq:f_divergence} as:
\begin{equation}
\begin{aligned}
    \nabla_\theta \mD_{\text{KL}}(p_\theta\parallel p_{\text{teacher}}) 
    &= \E_{\veps,\x,t}[-w(t)(\nabla_{\x_t}\log p_{\text{teacher}}(\x_t)-\nabla_{\x_t}\log p_{\theta^-}(\x_t))\nabla_\theta\mG_{\theta}(\veps)] \\
    &\approx \E_{\veps,\x,t}[-w(t)(\nabla_{\x_t}\log p_{\text{teacher}}(\x_t)-\nabla_{\x_t}\log p_{\phi}(\x_t))\nabla_\theta \mG_\theta(\veps)],
\end{aligned}
\label{eq:kl_divergence}
\end{equation}
where $\nabla_{\x_t}\log p_{\text{teacher}}(\x_t)$ and $\nabla_{\x_t}\log p_{\theta^-}(\x_t)$ represent the score functions for the teacher and student distributions. And $\phi$ represents the parameters of a fake score network that models the distribution of the students, as the student score score $\nabla_{\x_t}\log p_{\theta^-}(\x_t)$ is not intractable for the few-step generator $F_\theta$. 
Typically, the fake score network $F_\phi$ is initialized from the teacher model $\F_\text{teacher}$ with added LoRA~\cite{hu2022lora}, and only the LoRA parameters are trained. It uses the flow matching loss in \eqnref{eq:fm_objective} with data from the student model $\x\sim p_\theta$, thereby approximating the student distribution.

Leveraging $\nabla_{\x_t}\log p_\theta(\x_t)=-\frac{\x_t+(1-t)\F_\phi(\x_t,t)}{t}$ for flow matching~\cite{liu2023flow,song2020sde,ma2024sit,xu2025reals}, the gradient term in \eqnref{eq:kl_divergence} is:
\begin{equation}
\begin{aligned}
     \vg 
     &= -(\nabla_{\x_t}\log p^\text{CFG}_{\text{teacher}}(\x_t)-\nabla_{\x_t}\log p_{\phi}(\x_t)) \\
     &= \frac{1-t}{t}(\F^\text{CFG}_\text{teacher}(\x_t,t)-\F_\phi(\x_t,t)),
\end{aligned}
\label{eq:dm_gradient}
\end{equation}
where $\x_t=(1-t)\x+t\hat{\veps}$ with independently sampled noise $\hat{\veps}\sim \gN(\mathbf{0},\mathbf{I})$ and we use the teacher distribution $p^\text{CFG}_{\text{teacher}}(\x_t)$ with classifier-free guidance~\footnote{$p^\text{CFG}_{\text{teacher}}(\x_t)$ is represented by the teacher output with classifier-free guidance (CFG): $\F^\text{CFG}_\text{teacher}(\x_t,t)$.}.  
The distribution matching distillation loss is: 
\begin{equation}
    \gL_\theta^\text{DMD} = \frac{1}{2}\left\| \mG_\theta(\x_t,t) - \mG_{\theta^-}(\x_t,t) + \vg \right\|_2^2.
    \label{eq:dmd_loss}
\end{equation}



\paragraph{Summary.} Our framework culminates in a comprehensive framework for few-step learning.
We first train the network $\F_\theta$ with consistency loss \eqnref{eq:cm_loss} to obtain a few-step generator. Then we conduct a distribution matching distillation with \eqnref{eq:dmd_loss}, further improving the generation fidelity. 

\begin{table*}[!htbp]
    \centering
    \setlength{\tabcolsep}{1.0pt}
    \renewcommand{\arraystretch}{1.1}
    \scriptsize
    \caption{Performance on ImgEdit Bench and GEdit Bench.}
    \resizebox{\textwidth}{!}{
    \begin{tabular}{lccccccccccccc}
        \toprule
        \multirow{2}{*}{\textbf{Model}}
        & \multicolumn{10}{c}{\textbf{ImgEdit-Bench} $\uparrow$}
        & \multicolumn{3}{c}{\textbf{GEdit-Bench} $\uparrow$} \\
        & Extract & Style & Background & Add & Remove & Replace & Adjust & Compose & Action & Overall
        & G\_SC & G\_PQ & G\_O \\
        \midrule
        Qwen-Image-Edit~\cite{wu2025qwenimage}          &3.47 & 4.80 & 4.32 & 4.26 & 3.87 & 4.58 & 4.45 & 3.91 & 4.59 & 4.25 &8.18 & 7.87 & 7.68 \\
        Qwen-Image-Edit-2509~\cite{wu2025qwenimage}     &3.51 & 4.84 & 4.36 & 4.43 & 4.29 & 4.66 & 4.42 & 3.72 & 4.58 & 4.31& 8.12 & 8.01 & 7.61   \\
        Nano Banana~\cite{Nano-Banana} &3.89 & 4.2 & 4.32 & 4.33 & 4.39 & 4.55 & 4.36 & 3.42 & 4.48 & 4.22 & 7.43 & 8.14& 7.20\\
        Seedream 4.0~\cite{seedream2025seedream40nextgenerationmultimodal} & 2.96 & 4.76 & 4.22 & 4.47 & 4.25 & 4.42 & 4.31 & 3.11 & 4.45 & 4.11 & 8.24 & 7.86 & 7.66 \\
        UniPic 2.0~\cite{wei2025skywork} & 1.86 & 4.53 & 4.73 & 4.48 & 4.00 & 4.73 & 4.18 & 3.82 & 4.22 & 4.06 & 7.63 & 7.17 & 7.10 \\
        \midrule
        UniPic 3.0         &3.31 & 4.97 & 4.35 & 4.45 & 4.46 & 4.71 & 4.44 & 3.77 & 4.69 & 4.35 & 8.12 & 7.79 & 7.55  \\
        \bottomrule
    \end{tabular}
    }
    \label{tab:imgedit_gedit}
\end{table*}

\section{Experiments}
\subsection{Setup.}
The training data comprises 338K multi-image composition samples (including 215K internally constructed samples and Mico-150K~\cite{wei2025mico150kcomprehensivedatasetadvancing}) and 381k single-image editing samples from open-source datasets (Nano-consistent-150K~\cite{ye2025echo4o}, Pico-Banana-400K~\cite{qian2025picobanana400klargescaledatasettextguided}). We perform a full-parameter fine-tuning of the MMDiT model, training it for $80K$ steps with a global batch size of 64 and a learning rate of $1\times 10^{-4}$. After pre-training, we first conduct a consistency tuning for $10K$ steps with a global batch size of 256 and a learning rate of $1\times10^{-6}$. Based on the consistency model, we perform the distribution matching distillation for $10K$ steps with a global batch size of 64. We use the learning rate for the student model and the fake score network as $2\times10^{-6}$ and $4\times10^{-7}$, respectively. Teacher distribution uses the CFG scale as $w=6.0$. The training employs cosine learning rate annealing and the AdamW optimizer with $\beta_1 = 0.9$, $\beta_2 = 0.95$, $\epsilon = 1 \times 10^{-8}$, and a weight decay of $0.05$.

\subsection{Multi-Image Composition Benchmark.} Recognizing the absence of standardized evaluation protocols for multi-image composition, we construct \textbf{MultiCom-Bench}, a carefully curated benchmark comprising 200 high-quality triplets specifically targeting HOI scenarios. The benchmark is balanced between interaction types and input complexity (100 triplets with 2–3 source images, another 100 with 4–6 images). Following VIEScore~\cite{liu2025step1x}, we have designed stable and effective evaluation templates that assess model-generated results in multiple dimensions, including the adherence to composition instructions, image quality, and facial consistency. This benchmark will be released to facilitate future research in multi-image composition.

\begin{table}[h]
\centering
\small
\caption{Performance comparison of different models on multi-image composition.}
\label{tab:performance_comparison}
\begin{tabular}{lccc}
\toprule
\multirow{2}{*}{\textbf{Model}} & \multicolumn{3}{c}{\textbf{MultiCom-Bench}} \\
\cmidrule(lr){2-4}
 & 2-3 Images & 4-6 Images & Overall \\
\midrule
Qwen-Image-Edit~\cite{wu2025qwenimage} & 0.7705 & 0.4793 & 0.6249 \\
Qwen-Image-Edit-2509~\cite{wu2025qwenimage} & 0.8152 & 0.2474 & 0.5313 \\
Nano-Banana~\cite{Nano-Banana} & 0.7982 & 0.6466 & 0.7224 \\
Seedream 4.0~\cite{seedream2025seedream40nextgenerationmultimodal} & 0.7997 & 0.6197 & 0.7088 \\
\midrule
UniPic 3.0 & 0.8214 & 0.6296 & 0.7255 \\
\bottomrule
\end{tabular}
\end{table}

\begin{figure*}[htbp]
    \centering
    \makebox[\textwidth][c]{%
        \includegraphics[width=\dimexpr\paperwidth-0.5cm\relax]{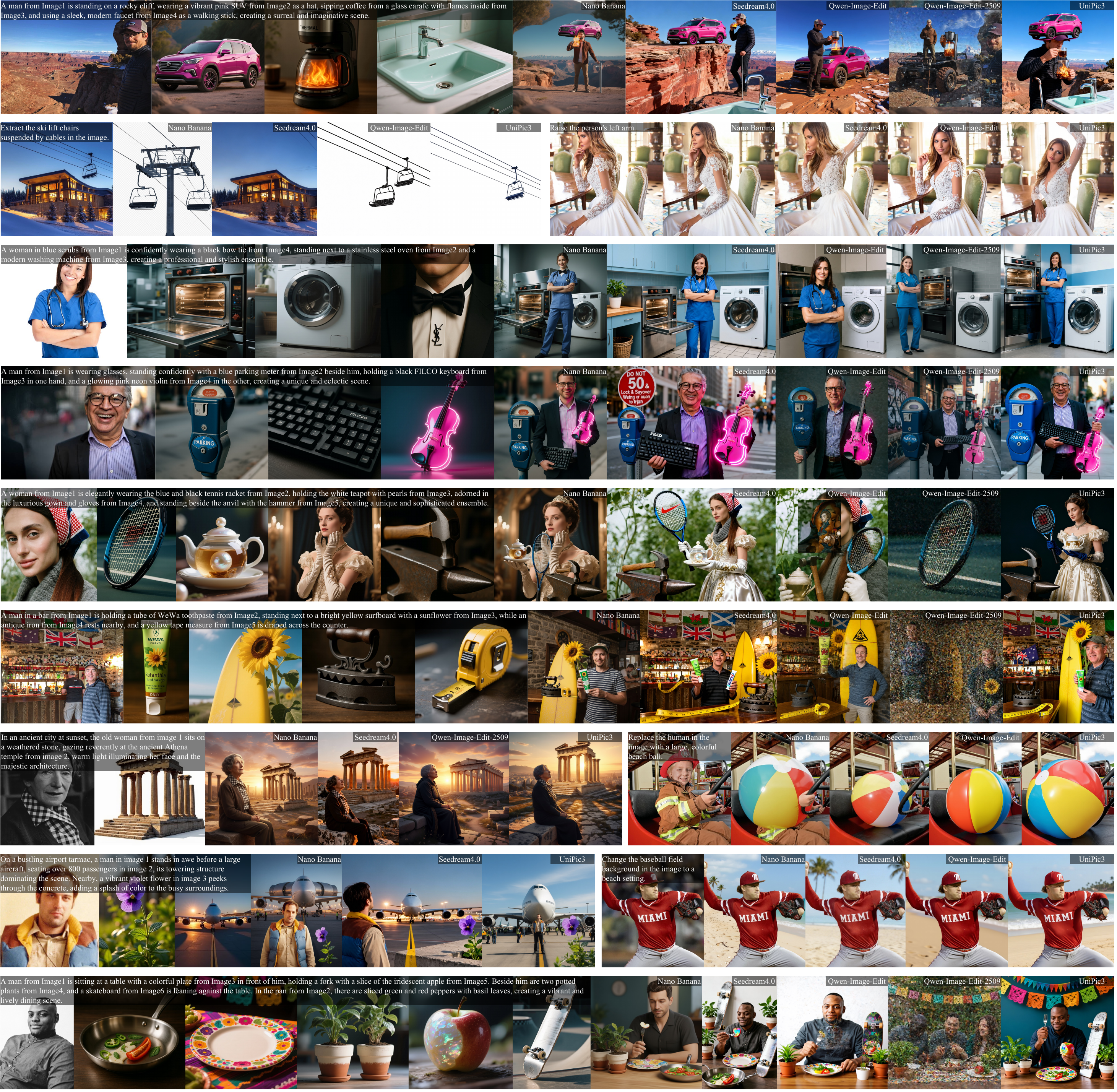}
    }
    \caption{Qualitative comparison among Nano-Banana, SeedDream4, Qwen-Image-Edit, Qwen-Image-Edit-2509, and UniPic 3.0. Our model demonstrates competitive or even superior performance in instruction-guided image editing and composition.}
    \label{fig:vis_edit}
\end{figure*}


\begin{figure}[htbp]
    \centering
    \includegraphics[width=1\linewidth]{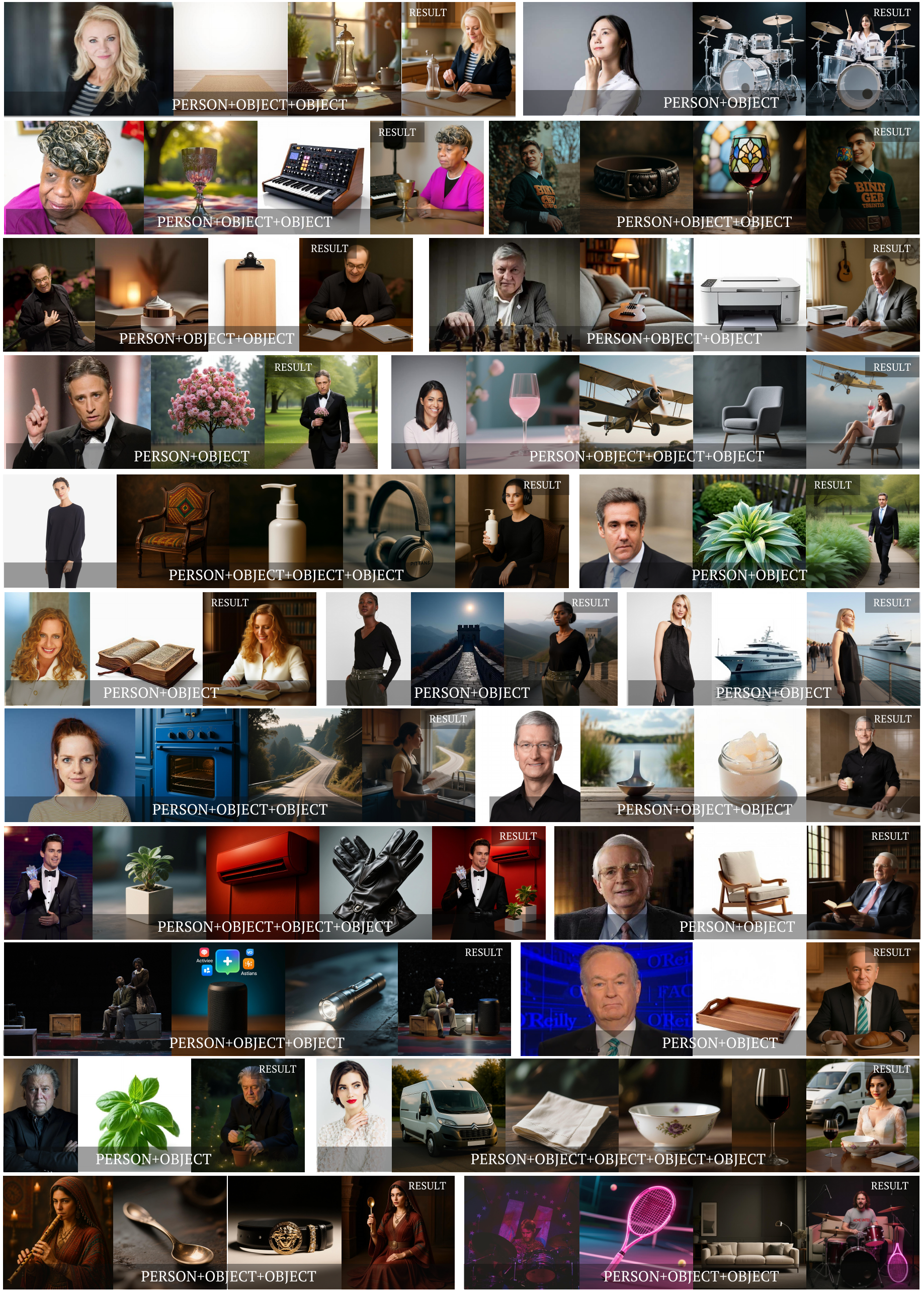}
    \caption{Qualitative results produced by our distilled UniPic 3.0 in just 8 steps.}
    \label{fig:few_step_demo}
\end{figure}

\subsection{Main results.} Table~\ref{tab:imgedit_gedit} and Table~\ref{tab:performance_comparison} summarizes our main comparative results, evaluating our models against other advanced methods on image editing, and multi-image composition. We compare with Qwen-Image-Edit~\cite{wu2025qwenimage}, Qwen-Image-Edit-2509~\cite{wu2025qwenimage}, UniPic 2.0~\cite{wei2025skywork}, Nano-Banana~\cite{Nano-Banana}, Seedream 4.0~\cite{seedream2025seedream40nextgenerationmultimodal}. The results clearly demonstrate the exceptional performance of our approach across major benchmarks, highlighting its powerful unified capabilities. Detailed comparisons for image editing and multi-image composition are provided in subsequent subsections.

\paragraph{Image Editing.}\label{subsec:edit}

We first evaluate UniPic 3.0 on single-image editing using two standard benchmarks: ImgEdit-Bench~\cite{ye2025imgedit} and GEdit-Bench~\cite{liu2025step1x}. As shown in Table~\ref{tab:imgedit_gedit}, UniPic 3.0 achieves overall scores of 4.35 on ImgEdit-Bench and 7.55 on GEdit-Bench. These results indicate that UniPic 3.0 preserves strong single-image editing performance while being equipped with unified multi-image composition capabilities.
\paragraph{Multi-image Composition.}\label{subsec:multi_image_composition}
As shown in Table~\ref{tab:performance_comparison}, we compare UniPic 3.0 with state-of-the-art methods, including Qwen-Image-Edit, Nano-Banana, and Seedream 4.0. It is worth noting that Qwen-Image-Edit does not natively support multi-image composition. However, through our proposed inference paradigm, the model exhibits emergent capabilities for this task. Nevertheless, this capability is limited to the 2-3 image setting, as the performance degrades significantly when extending to 4-6 images. Regarding the Qwen-Image-Edit-2509 model, the developers claim that optimal performance is achieved with 2-3 input images. Our evaluation using the official inference code shows that the model performs well on the "2-3 Images" subset, where it is second only to UniPic 3.0. However, its performance degrades significantly on the "4-6 Images" subset.

Our model achieves the best performance with an Overall score of 0.7255, outperforming the second-best method, Nano-Banana, by a recognizable margin. Specifically, UniPic 3.0 demonstrates exceptional capability in scenarios with fewer inputs, reaching a score of 0.8214 in the "2-3 Images" subset. This significantly surpasses Seedream 4.0 (0.7997) and Nano-Banana (0.7982), indicating our model's superior precision in handling basic composition tasks. 



\paragraph{Qualitative Results.} Figure \ref{fig:vis_edit} demonstrates a qualitative comparison with the leading models in multi-image composition. Consistent with the quantitative results, UniPic 3.0 can provide more precise control and show enhanced instruction-following capability. In Figure~\ref{fig:few_step_demo}, we provide the visualization results for the generation in a few-steps. In particular, our distilled UniPic 3.0 can produce high-fidelity results in just 8 steps.

\section{Conclusion}
In this work, we presented Skywork UniPic 3.0, a unified framework that bridges the gap between single-image editing and complex multi-image composition. By formulating these distinct tasks as a conditional generation problem on a unified sequence, we eliminate the need for task-specific architectures and enable the model to handle arbitrary inputs and flexible resolutions. Our extensive empirical analysis highlights two critical findings: first, that a rigorously curated dataset prioritizing semantic coherence, specifically for Human-Object Interaction (HOI), is more effective than massive, noisy datasets; and second, that integrating trajectory mapping with distribution matching significantly accelerates inference without compromising fidelity.

Skywork UniPic 3.0 demonstrates state-of-the-art performance on established single-image editing benchmarks and sets a new standard on the proposed MultiCom-Bench, surpassing leading commercial baselines like Nano-Banana and Seedream 4.0. We believe this work not only provides a reproducible methodology for high-quality multi-image composition but also establishes a strong baseline for future research into unified, efficient, and interactive generative systems.

\section{Contributors}
Hongyang Wei$^*$, Hongbo Liu$^*$, Zidong Wang$^*$, Yi Peng$^*$, Baixin Xu, Size Wu, Xuying Zhang, Xianglong He, Zexiang Liu, Peiyu Wang, Xuchen Song$^{\dag}$, Yangguang Li$^{\dag}$, Yang Liu, Yahui Zhou\\

$^*$ Equal contribution.\\
$^{\dag}$ Corresponding author.

\clearpage
\bibliography{main}
\bibliographystyle{plain}
\end{document}